\documentclass{article}

\usepackage{PRIMEarxiv}

\usepackage[utf8]{inputenc} 
\usepackage[T1]{fontenc}    
\usepackage{hyperref}       
\usepackage{url}            
\usepackage{booktabs}       
\usepackage{amsfonts}       
\usepackage{nicefrac}       
\usepackage{microtype}      
\usepackage{lipsum}
\usepackage{fancyhdr}       
\usepackage{graphicx}       
\usepackage{subfig}
\graphicspath{{media/}}     
\DeclareUnicodeCharacter{00A0}{ }

\usepackage{xcolor}
\definecolor{verde}{rgb}{0.25,0.5,0.35}
\definecolor{jpurple}{rgb}{0.5,0,0.35}
\definecolor{darkgreen}{rgb}{0.0, 0.2, 0.13}
\usepackage{listings}

\newcommand{\estiloPython}{
  \lstset{ %
    language=Python,                     
    basicstyle=\footnotesize,       
    numbers=left,                   
    numberstyle=\tiny\color{gray},  
    stepnumber=1,                   
    numbersep=5pt,                  
    backgroundcolor=\color{white},  
    showspaces=false,               
    showstringspaces=false,         
    showtabs=false,                 
    frame=single,                   
    rulecolor=\color{black},        
    tabsize=2,                      
    captionpos=b,                   
    breaklines=true,                
    breakatwhitespace=false,        
    title=\lstname,                 
    keywordstyle=\color{blue},      
    commentstyle=\color{darkgreen},   
    stringstyle=\color{red},      
    escapeinside={\%*}{*)},         
    morekeywords={*,...}          
}}

\title{The use of Data Augmentation as a technique for improving neural network accuracy in detecting fake news about COVID-19}

\author{
  Wilton O. Júnior, Mauricio S. da Cruz, André Brasil Vieira Wyzykowski, Arnaldo Bispo de Jesus \\
  Instituto de Informática -- Universidade Católica do Salvador (UCSAL) \\
  Salvador - Bahia - Brazil\\
  \texttt{\{wilton.junior, mauricio.cruz\}ucsal.edu.br} \\

}

\begin{document}
\maketitle

\begin{abstract}
  This paper aims to present how the application of Natural Language Processing (NLP) and data augmentation techniques can improve the performance of a neural network for better detection of fake news in the Portuguese language. Fake news is one of the main controversies during the growth of the internet in the last decade. Verifying what is fact and what is false has proven to be a difficult task, while the dissemination of false news is much faster, which leads to the need for the creation of tools that, automated, assist in the process of verification of what is fact and what is false. In order to bring a solution, an experiment was developed with neural network using news, real and fake, which were never seen by artificial intelligence (AI). There was a significant performance in the news classification after the application of the mentioned techniques.
\end{abstract}


\section{Introduction}

The COVID-19 pandemic emerged within a context in which people are informed more by headlines on social media like Facebook than by the content of the newspaper article, opening up gaps for the phenomenon of fake news. The discovery of a new virus demands news dealing with the symptoms, severity, and forms of transmission. Furthermore, as everything is new and volatile, it is a fertile ground for the insertion of disinformation, especially in a scenario in which the habit of checking the veracity of information does not happen often. The production of fake news happens at a speed that is impossible to be checked by a small group of journalists, and these professionals must also cover the news coming from reliable sources. Given this context, the automation of this activity seems to be the only effective way to combat fake news through Artificial Intelligence (AI). 

It is worth mentioning that the fight against fake news has a huge importance and responsibility of determining something as being false or true. Given that, there is the ability to innovate in producing new frauds. That's why works like this and others mentioned in this article have a lot of relevance in this pandemic context.

Several authors such as \cite{madani2021using}, \cite{patwa2020fighting} and \cite{mookdarsanit2021covid} have already proposed algorithms, many using modern AI techniques such as Neural Processing Language (NLP) or multi-layered Perceptron (MLP). Most of this work is related to \textit{contenders} promoted on the Kaggle platform\footnote{Kaggle: an online community of data scientists and machine learning professionals subsidiary of Google.}. Some authors like \cite{ding2020challenges} and \cite{patwa2020fighting} use the term \textit{infodemic} to describe this phenomenon of fake news about the coronavirus.

Some social networks, like Twitter and Facebook, have released tools to combat fake news. In the Facebook's case, a series of authentications to ensure that the person responsible for the post is the real owner of the account \cite{FacebookFakeNewsTool} and Twitter's Birdwatch\footnote{Source: https://tinyurl.com/ferramentamessenger}, a tool that allows the user to report posts as fake news. However, there are several reasons to suspect the interest of these corporations in combating this practice since, when verifying the source, the user leaves the platform, in contrast with the business model of a social network that focus on engagement. 

\section{Background}
With the advance of the COVID-19 pandemic, the discussion of whether access to internet is one of the fundamental rights for survival (as well as water, electricity, and basic sanitation) is no longer controversial and has become a consensus, given the fact that the dynamics of modern society demands that, for a citizen to be included, he must be connected. \cite{internetEducacao}, talks about how the democratization of the internet is fundamental for the maintenance of the basic right to education provided in the Brazilian constitution of 1988. According to the Collins Dictionary, \textit{lockdown} is the word of the year for 2020, considering that the dictionary recorded around 250 thousand uses against only 4 thousand the previous year \cite{Collins2020}.

For many years, the media's credibility, so be it print, radio or television, was practically irrefutable. Today, with the democratization of access, and especially the production of information and content, new vehicles, websites, \textit{blogs}, YouTube channels, and other platforms have also started to act as press \cite{kalsnes2018fake}. \cite{kalsnes2018fake} goes on to say that, even after the fake news is denied, it still influences the population, as it attacks not only the facts but also the credibility of important institutions such as press vehicles. In addition, social networks have a tremendous impact on the way people think and show themselves as a powerful weapon of manipulation, leading political groups to create fake news or spread half-truths in order to create or influence a movement. 

In 2020, some authors analyzed the impact and danger of fake news during the pandemic. Kalsnes cites in his work that the main reasons behind fake news include political, financial, and social issues \cite{kalsnes2018fake}. \cite{neto2020fake} Shows how a series of fake news can be at the service of an ideology such as the privatization of the Brazilian Unified Health System (Sistema Único de Saúde - SUS) since these false articles try to discredit an organ of such tradition and importance.

Still, in his analysis \cite{neto2020fake}, says that fake news in the Brazilian context about COVID-19 is divided into five categories. They are information related to health authorities, therapy, prevention measures, disease prognosis, and vaccination. These are the five main points, and their primary vehicles are social networks, especially \textit{WhatsApp}.

The power of fake news has been extremely underestimated in the past. From the 2016 elections in the United States to the elections in Brazil in 2018 and now during the coronavirus pandemic, where President Bolsonaro accused press vehicles like Globo Television, and the Folha de SP newspaper in 2018 \cite{pereira2021desinformaccao}. This behavior is similar to that of the President of the United States, Donald Trump, whom several times classified information coming from CNN and \textit{New York Times} as fake news \cite{kalsnes2018fake}.

Efforts to combat fake news have proved quite challenging for traditional methods, as producing fake news takes less time and effort than debunking it. One of these efforts is Lupa, a project carried out by the Folha de São Paulo newspaper, which is a \textit{Fact-Checking} agency, the first in Brazil \cite{Lupa2020}. However, it has been observed that there is a pattern in fake news created by the context of the pandemic, and through it, it's possible to automate this validation.

\section{Related Works}
Several researchers worldwide have been dedicated to creating solutions for the detection of fake news. However, there is still a lot of debate regarding the best architecture for this purpose. \cite{singh2020fake} He did a work comparing several techniques of modern artificial intelligence such as Convolutional Neural Network (CNN), Artificial Neural Network (ANN), and Recurrent Neural Network (RNN), and the accuracy varied a lot according to the \textit{dataset} and the preprocessing the data before training.

The fact that it is a classification, it is necessary to portray the similarity of the data through a vector representation. This consists of analyzing the curve between the expected results with the results obtained and thus obtaining the accuracy and the \textit{LOSS}, that is, how much is lost, of a training.

\cite{singh2020fake} in his work, he made a comparison between the techniques \textit{Word2Vet}, \textit{One-Hot Encoding}, \textit{Doc2Vec} and TF-IDF, used text \textit{vectorization}, used based on the CNN, ANN and RNN architectures. He concluded that the data do not vary much according to the architecture, but it does trough the \textit{dataset}, which when using the TF-IDF technique with the \textit{dataset} Kaggle were obtained the best results in the range of 0.96.


In 2021, \cite{shushkevich2021tudublin} participated in the \textit{TUDublin team at Constraint@AAAI2021 - COVID-19 Fake News Detection}, a competition where researchers with the same \textit{dataset} seek to obtain the best possible accuracy. In 2021, the theme was precisely the pandemic of the new coronavirus. At the end of the project, they obtained between 0.91 and 0.95 accuracy, which are encouraging results given the urgency of this category of work.

For this, the researchers analyzed the most frequent terms using TF-IDF and removed the most common affixes, morphological and inflectional endings to keep only the central idea of the word with the \textit{PorterStemmer} algorithm and, finally, changed the entire text to lowercase. After pre-processing, the model used the following architecture: \textit{logistic regression}\footnote{Logistic Regression: a statistical technique that aims to produce a model that allows the prediction of values taken by a categorical variable.}, SVM\ footnote{SVM: concept in computer science for a set of supervised learning methods used for classification and regression analysis.}, \textit{Naive Bayes}\footnote{Naive Bayes: family of simple "probabilistic classifiers" based on the application of the Bayes' theorem.} and a combination of \textit{Naive Bayer} with \textit{logistic regression}.

In another related work, researchers at Accenture, \cite{paka2021cross} created the \textit{Cross-SEAN}, which is based on the analysis of messages coming from social media, especially Twitter. For this, they divided the process into four stages. The first was to separate the \textit{tweets} related to the coronavirus, based on the posts available in the \textit{dataset} \textit{Kaggle}, the IDs of these \textit{tweets} were taken and searched in the API\footnote{API : set of routines and programming patterns for accessing Twitter's own \textit{software} or web-based platform.}. Then they collected reliable information about the pandemic using existing fact-checking tools (\textit{Snopes, PolitiFact, FactCheck, and TruthOrFiction}). From there, they extract URLs\footnote{URL: virtual address of a page or website.} with topics related to COVID-19, classifying the \textit{tweets} between genuine and fake regarding the pandemic using text manipulation techniques such as BERT \footnote{BERT model: a technique used for pre-training and natural language processing developed by Google for machine learning.} and ROBERTa\footnote{RoBERTa: pre-training model based on highly optimized BERT}. The last step performed was manually validating a sample of the results obtained. After the entire process, accuracy of 92\% was found among the false posts detected automatically.

Despite excellent initiatives, we miss in all these works their applicability in the news, since it is through them that public opinion is based, given the fact of trusting the credibility of the press agency.

\section{Materials and methods}
This research was adapted based on a work written by another author \cite{singularity014}. His repository can be found in the references of this article. Furthermore, we propose to increase the accuracy of this algorithm by applying \textit{data augmentation} techniques to texts, in addition to translating the \textit{dataset} in order to try to bring the research application to the national territory due to the scarcity of datasets to be used on the proposed topic.

Although there are good works such as Fake Br Corpus \cite{Fake.Br} and \cite{dias2019towards} in Brazilian Portuguese for fake news. They do not address the context of the new coronavirus pandemic. There is a gap between the works available in Portuguese and the information on COVID-19, and this ends up becoming a challenge in this area. The only alternative found was LATAM Coronavirus \cite{LATAMcoronavirus}, a constantly updated database used to check, validate and explain information about the new coronavirus. However, this database is not well-structured for news analysis. All the code is available on Google Colab\footnote{Google Colab: Colaboratory or "Colab" allows you to write Python code in your browser without any necessary configuration and with free access to GPUs.}. The link is in the references of this article, as well as the link to GitHub\footnote{GitHub: hosting platform for source code and files that allows any user registered on the platform to contribute to private and Open Source projects.} with the author's project, which we use as a base.

\section{Technologies used}

In this topic, we will cover all the technologies applied during the development process of this work.

\subsection{\textit{Python}}Launched by Guido Van Rossum in 1991, Python is a high-level programming language used for object-oriented programming (OOP) and structured programming. The language was designed with the philosophy of emphasizing the importance of programmer effort over \cite{python} computational effort.

There are several reasons why data scientists use Python. These need to create data visualizations to communicate results and predictions at any business level. With that in mind, the Python language has a great advantage because it contains libraries, \textit{frameworks}\footnote{Framework: a set of ready-made codes that can be used in application development. } and exclusive packages to be used in the data area, such as \textit{sci-kit learn} for \textit{Machine Learning}, \textit{Numpy}, \textit{Pandas} for data analysis, among others \cite {datascientist}.

\subsection{\textit{Dataset}}
 \textit{Dataset} is the term used for a collection of tabulated information (data), where each column represents a variable, and the particular value of each row corresponds to a dataset in question.

\textit{datasets} can contain information, such as medical records or insurance records, for use by a program running on the system. These are also used to store information needed by applications or the operating system itself, such as source programs and macro libraries\footnote{Macro: sequences of programmed events (such as keystrokes, clicks, and delays) that assist with repetitive tasks.} or system variables or parameters. These can be cataloged, which allows the \textit{dataset} to be referred to by name without specifying where \cite{ibmdataset} is stored.

\subsection{\textit{Natural Language Processing} (NLP)}
It is a subfield of Artificial Intelligence that gives the computer the ability to understand, analyze, manipulate and reproduce human speech/writing with considerable accuracy. It is also used to study and understand natural human language in its entirety \cite{NLP}.

Currently, Google's search engine uses NLP to find and return results similar to the keywords entered the search field. In addition, it uses NLP in Google Translate, making the translation from one language to another much more efficient, not letting the text lose the sense of context \cite{googlenlp}.

\subsection{Tokenization}
Tokenization is the technique used to break the sequence of characters in a text by locating the limit of each word, separating each term of a sentence into tokens, a term used for computational purposes. Tokenization can normalize a text, for example, by mapping words to lowercase-only versions, expanding contractions, and even extracting the root of each word \cite{nltk}.

\subsection{\textit{SpaCy}}
SpaCy is an open-source library used in the Python programming language for NLP (Natural Language Processing). This library already has pre-built models that act as ``trained brains'~ for each language, making it easy to use for various purposes \cite{SpaCy}.

In addition to being used in NLP, this library contains processing tools \textit{POS-Tagging} (\textit{Part-Of-Speeching Tagging} (or understood as ``Grammar Analysis") for detecting adjectives, nouns, among others. Grammatical classes in the text, as well as being used as NER (\textit{Named Entity Recognition}) to determine if a term refers to a person, date, place, organization, among others, in order to improve the identification accuracy of correct terms.

\subsection{PyDictionary}
PyDictionary is a library for getting meanings, translations, synonyms, and antonyms. It uses WordNet for meanings, Google for translations, and synonym.com for synonyms and antonyms \cite{pydictionary}.

\subsection{Neural Network}These are computational models inspired by neurons in the human brain. With proper training, they can recognize patterns and correlations in raw data, be able to group and classify them, and over time, be able to learn continuously. Its first appearance was in 1943 by Warren McCulloch and Walter Pitts. From an article created on how neurons worked, they modeled a simple neural network using \cite{neuralnetowrk} electronic circuits.

\subsection{Data Augmentation}
The data augmentation technique consists of manipulating the \textit{dataset} in order to create similar data for training in order to increase the amount of data, adding slightly modified copies of existing data or newly created synthetic data from existing data. However, for a \textit{data augmentation} to be efficient, the manipulation cannot mischaracterize the \textit{dataset}. In other words: a recommendation from the World Health Organization (WHO) must still be true and correspond with the context of COVID-19 \cite{tensorflow}.

\section{Development}
The critical point of this research is to improve a set of pre-formatted data in order to increase the amount of information contained in it through the \textit{data augmentation} technique so that, after training, the neural network has a percentage of accuracy superior to the previous one, increasing its effectiveness. However, there is still another factor that we need to consider: the \textit{dataset} chosen is in the English language. That means the whole approach would have to be done by obeying the grammatical rules of the language. There was an attempt to search for a \textit{dataset} in the native language (Portuguese). However, Brazil lacks a set of pre-defined data to be freely analyzed in this context. Because of this, the choice of the foreign \textit{dataset} was the fastest and closest option found for the research development.

In total, 04 \textit{datasets} were separated: one for the real training (\textit{Real Train}), one containing real news that will be used for testing (\textit{Real Test}), one for the training \textit{fake} (\textit{Fake Train}) and the last one containing fake news (\textit{Fake Test}), without having to be used in that order, where all \textit{datasets} ``\textit{Train }"~has 800 news and \textit{datasets} ``\textit{Test}"~has 200.

The challenge was to translate all the \textit{dataset} into Portuguese in order to keep the entire grammatical structure intelligible so that the neural network could maintain accuracy or have the least possible loss because the validation will occur with national (real or false), translation from one language to another can significantly impact the accuracy of neural network analysis. For this, the library \textit{Google Translate} was used to do the translation properly. The translation process took 36 hours to complete. This factor was aggravated because the chosen library did not translate texts that contained more than 5,000 (five thousand) characters. Which led to the following specific instructions:

\begin{enumerate}
\item Create a loop to loop through the entire \textit{dataset}
\item Check if the current text is longer than 5000 characters
\item If yes, we translate the text sentence by sentence, detecting through a period (.) in the text.
\item If not, the library translates all unprocessed text.
\end{enumerate}

\begin{scriptsize}
\estiloPython
\begin{lstlisting}[caption={Source code in Python.} {Source: The Author}, label=lst:rcode]
def translate_text(word):
  translator = google_translator()
  tword = translator.translate(word, lang_src=``en",lang_tgt=``pt")
  time.sleep(1)
  return tword
\end{lstlisting}

\end{scriptsize}

After the translation process carried out by the function shown in Listing I, it was necessary to visualize the most common words in real or fake news, based on the \textit{datasets} used, to discover any pattern that differentiated them. For that, the library \textit{WordCloud} was used, efficient to visualize the most used words in both situations. A \textit{word cloud} was generated for the \textit{dataset} translated ``\textit{Real Train}"~and another \textit{word cloud} for the \textit{dataset} translated ``\textit{Fake Train }". The \textit{datasets} ``\textit{Test}"~ are for validation purposes only.

\begin{figure}[hbt!]
     \centering
     \subfloat[][Most common words in real news]{\includegraphics[width=.5\textwidth]{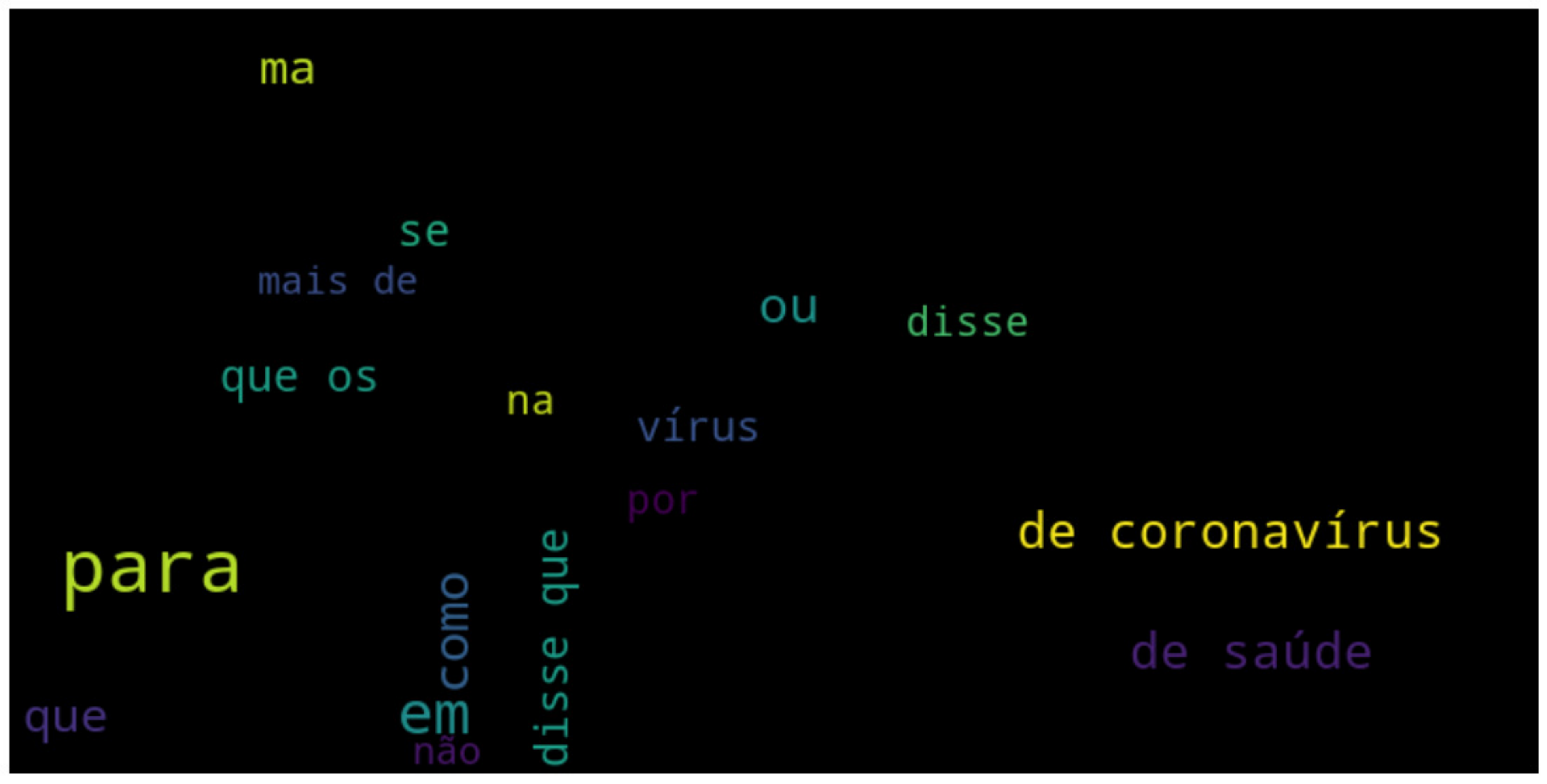}}
     \subfloat[][Most common words in fake news]{\includegraphics[width=.5\textwidth]{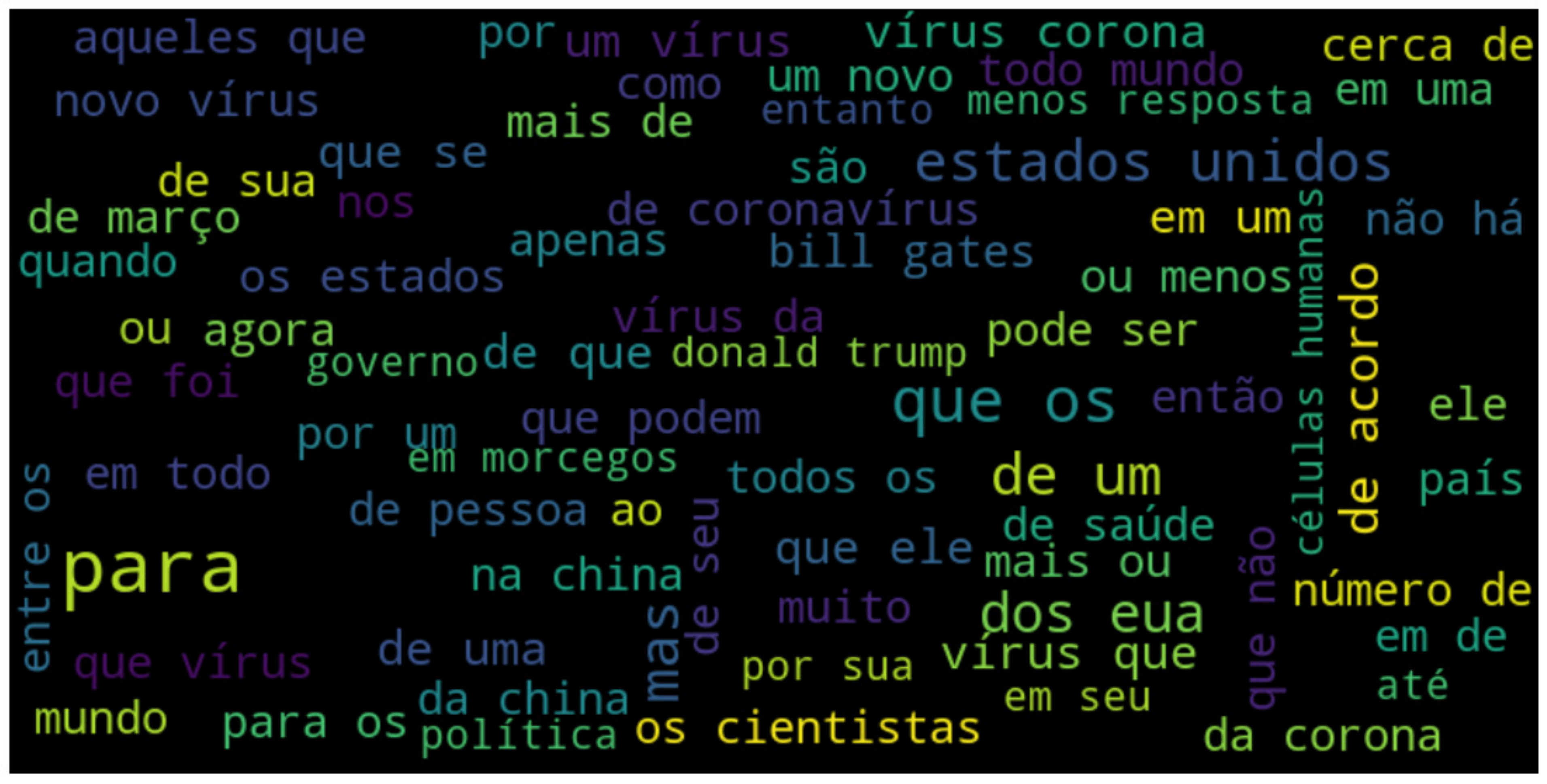}}
     \caption{\textit{Word clouds} after processing.} {Source: The author}
     \label{steady_state}
\end{figure}

Figure 1(a) indicates the most common words in real news based on the \textit{dataset} ``\textit{Real Train}". However, it shows a critical flaw: when translating the \textit{dataset}, the \textit{word cloud} appears to have a lack of words which, considering the size of the chosen dataset, does not reflect the ideal amount of words, while the \textit{word cloud} shown in Figure 1 (b), from \textit{ dataset} ``\textit{"Fake Train}", has a much higher consistency of words. It is noticed that, when the texts are translated, the words with neutral contexts of the Portuguese language tend to stand out as the most used words in real news. In the case of \textit{dataset} \textit{fake}, the words in common tend to have more diversities, such as proper names, countries, and political contexts. In short, what can be highlighted is:
\begin{enumerate}
\item The \textit{word cloud} of the \textit{dataset} ``\textit{Real Train}"~indicates that the most common words are those with neutral contexts (``to", ``that", ``as ", ``more than"...)
\item The \textit{word cloud} of the \textit{dataset} ``\textit{Fake Train}"~already covers contextualized keywords, such as ``Bill Gates", ``Donald Trump", ``Bats", ``human cells", among other words.
\end{enumerate}

However, it is necessary to have in mind that the \textit{word cloud} of the \textit{dataset} \textit{``Real Train"} maintains a false representation of the dataset used. With the size of both datasets, the figures would need to present several words as close to each other as possible. However, Figure 1 (a) shows a huge lack of words compared to Figure 1 (b).

\begin{figure}[hbt!]
\centering
\includegraphics[width=.8\textwidth]{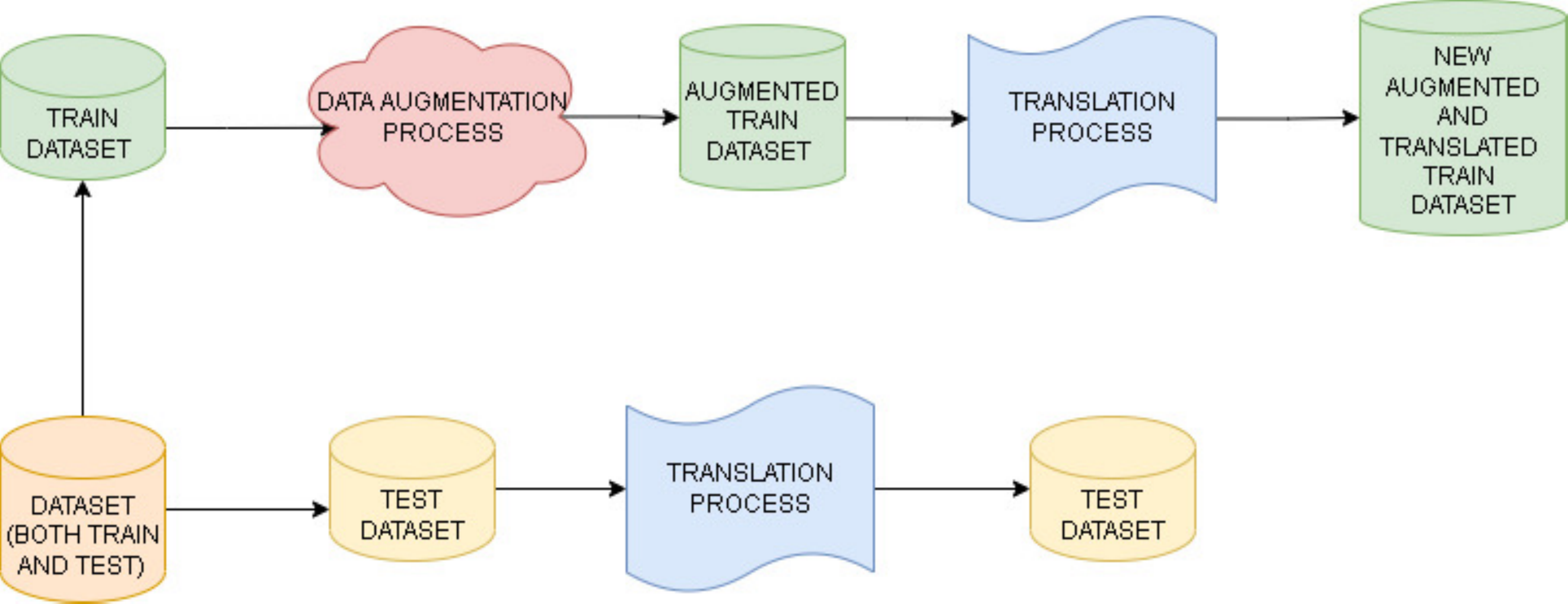}
\caption{Process of \textit{data augmentation} and translation of \textit{dataset}.}{Source: The Author}
\label{Function snippet}
\end{figure}

\begin{figure}[hbt!]
\centering
\includegraphics[width=.5\textwidth]{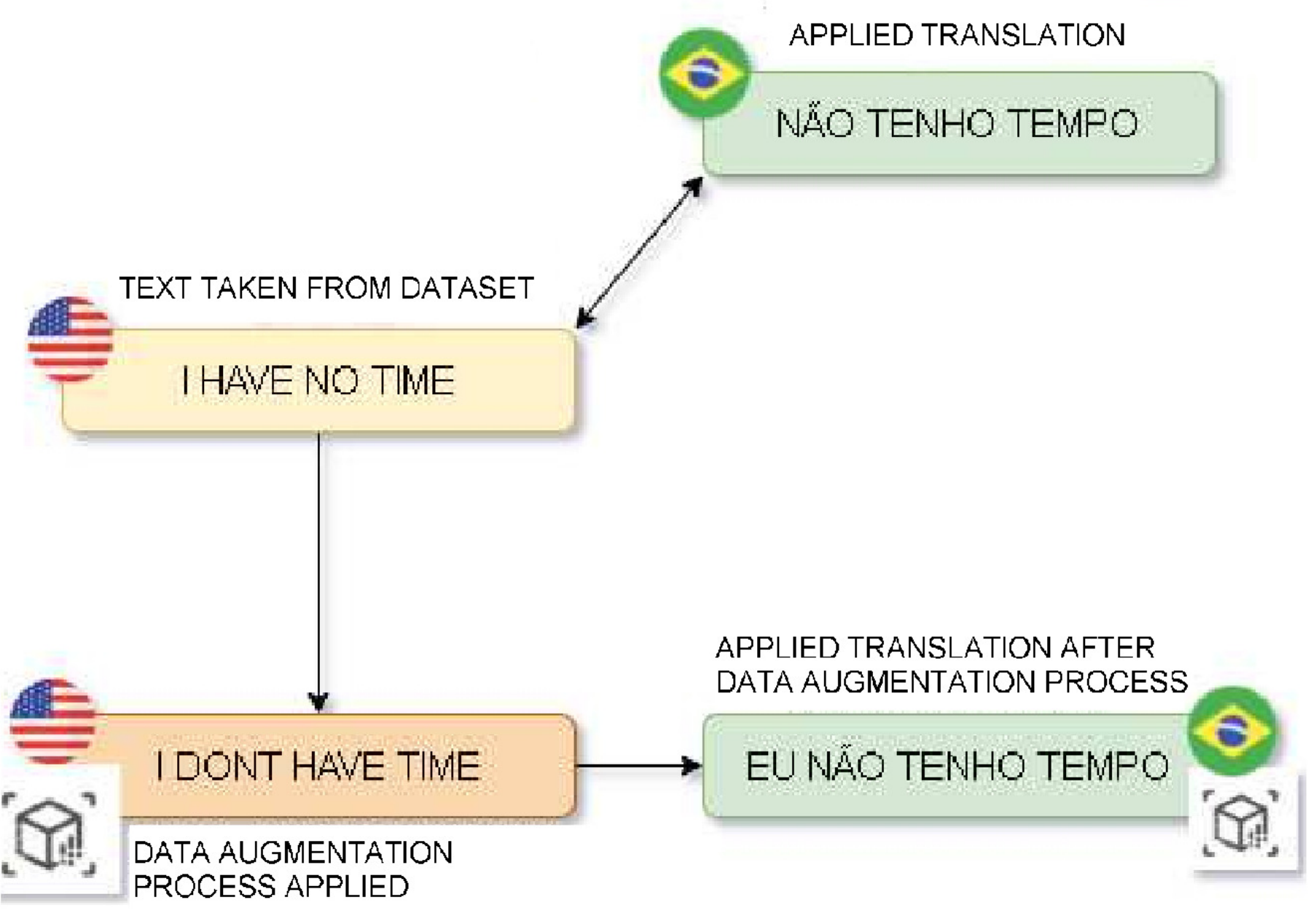}
\caption{Example of text translated without \textit{data augmentation} and translated with the technique applied.} {Source: The Author}
\label{}
\end{figure}
With that in mind, a process was built, as shown in Figure 2, where \textit{data augmentation} is applied even before the translation takes place. With this process, both \textit{datasets} would be much more abundant and very contextualized. To apply this technique, two libraries will be used: \textit{SpaCy} and \textit{PyDictionary}, to apply the NLP and search for synonyms of the chosen words, respectively. This example can also be seen in Figure 3, which shows how a sentence would look if translated in its entirety and how the same sentence would look if we applied \textit{data augmentation} first.

To not misdirects the news, we applied the first rule that all words exchanged would be nouns. This is because verbs, pronouns, and adjectives can quickly go out of context if synonyms replace them.

The second rule is that the noun is only changed if at least 40 percent similar to the chosen synonym. This number was taken based on the parameters of the \textit{SpaCy} library, known as the ``similarity index". Although this number may seem low, the library is quite demanding about its similarity. If the value is too high, the \textit{data augmentation} will not be robust, so it will have no impact during training.

\begin{figure}[hbt!]
\centering
\includegraphics[width=.7\textwidth]{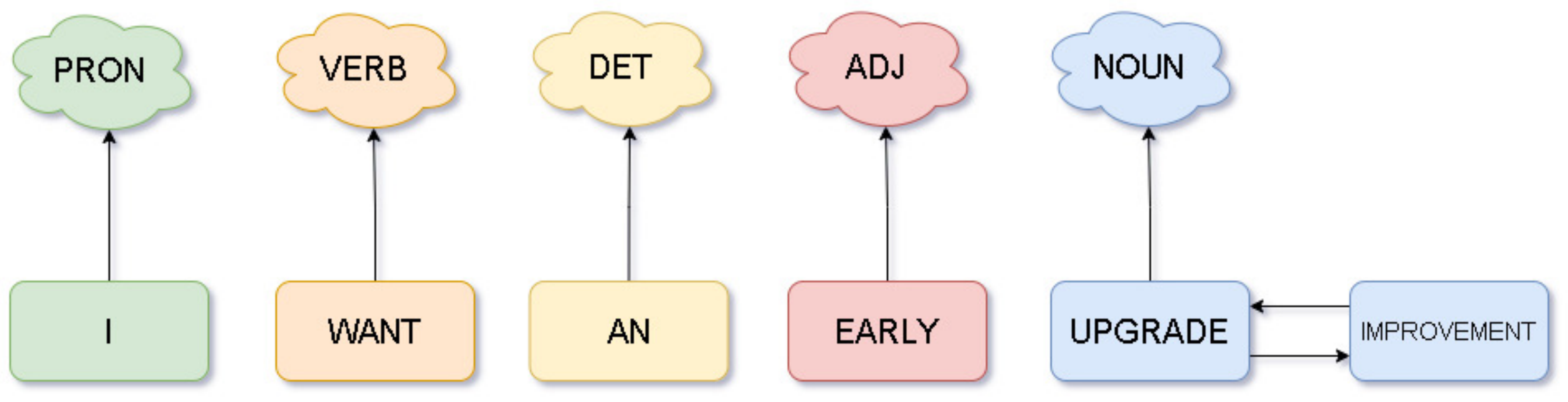}
\caption{Example of how the tokenization process, detection of grammatical classes and word change by synonym is done.}{Source: The Author}
\label{}
\end{figure}

For the first rule, the POS-Tagging (Part Of Speech Tagging) technique was used, as shown in the example of Figure 4, where the library had already a pre-trained model to be used, separating each word into a token\footnote{Token: Word of a text separated in an independent unit.} to be parsed in a unitary way. The POS-Tagging technique identifies which grammatical class that word belongs to (pronoun, noun, verb, adjective, among others). Only words in the ``noun" class will be identified.

For the second rule, we apply the PyDictionary library to search for synonyms for the tokenized noun, as shown in Figure 4. As this library can return a list containing several nouns, we use the similarity index from the \textit{SpaCy's} \footnote{SpaCy: spacy.io} library depending on the word to identify if it has a similarity above 40 percent to the token, replacing it with the synonym with the highest percentage if so. There is also the possibility that a noun has no synonyms. In this case, we continue without changing the word.

This process is repeated for each news item in a \textit{dataset}. In all, the process took about 60 hours to complete.
After completing the \textit{data augmentation} on the \textit{datasets} ``\textit{Train}" and the translation process, it was generated a new \textit{word cloud} of both \textit{datasets} (\textit{Real} and \textit{Fake}) to see if there has been any significant change in the most common words for each.
\begin{figure}[hbt!]
      \centering
      \subfloat[][Most common words in real news after \textit{data augmentation.}]{\includegraphics[width=.48\textwidth]{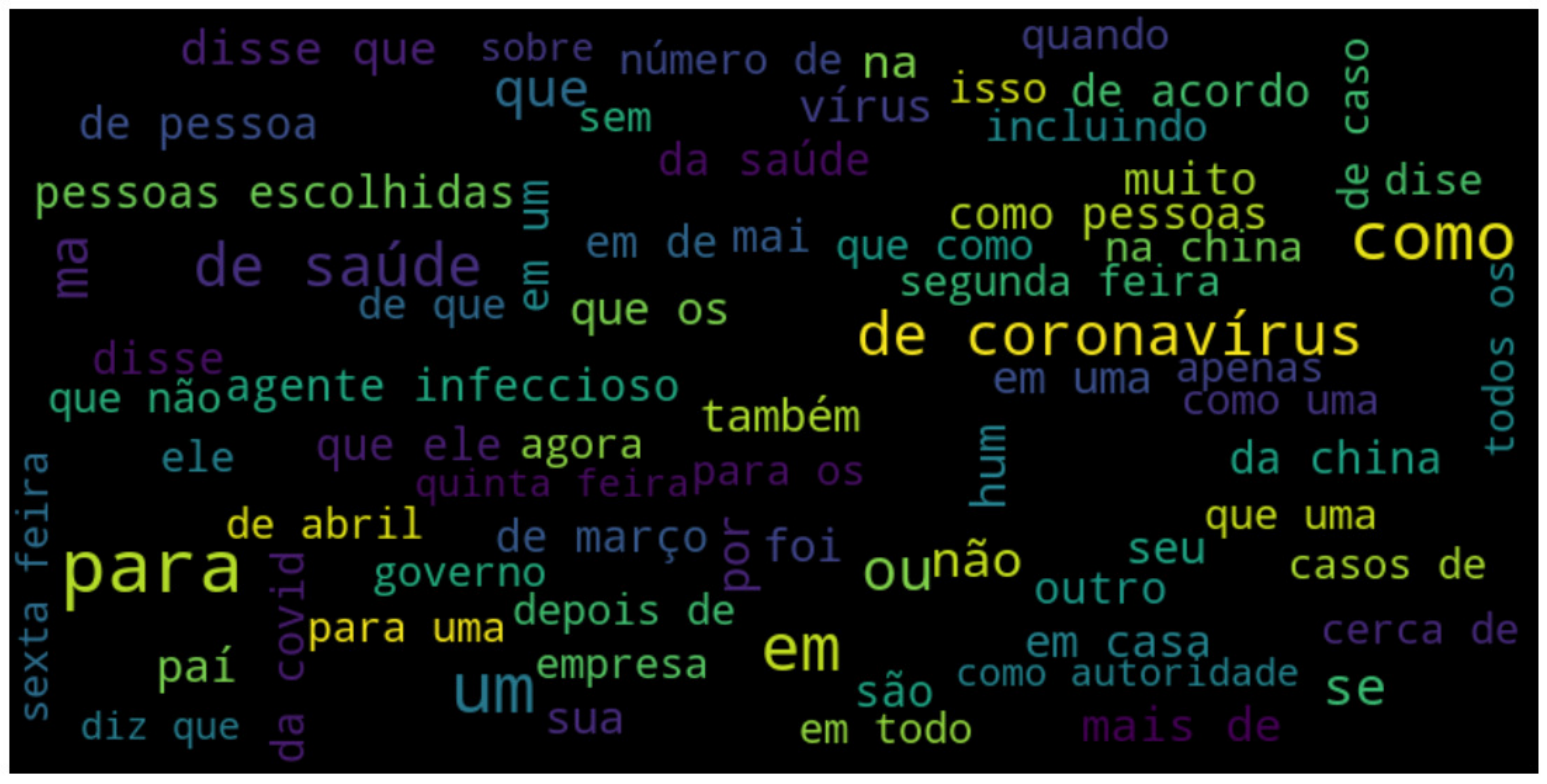}}
      \subfloat[][Most common words in fake news after \textit{data augmentation.}]{\includegraphics[width=.48\textwidth]{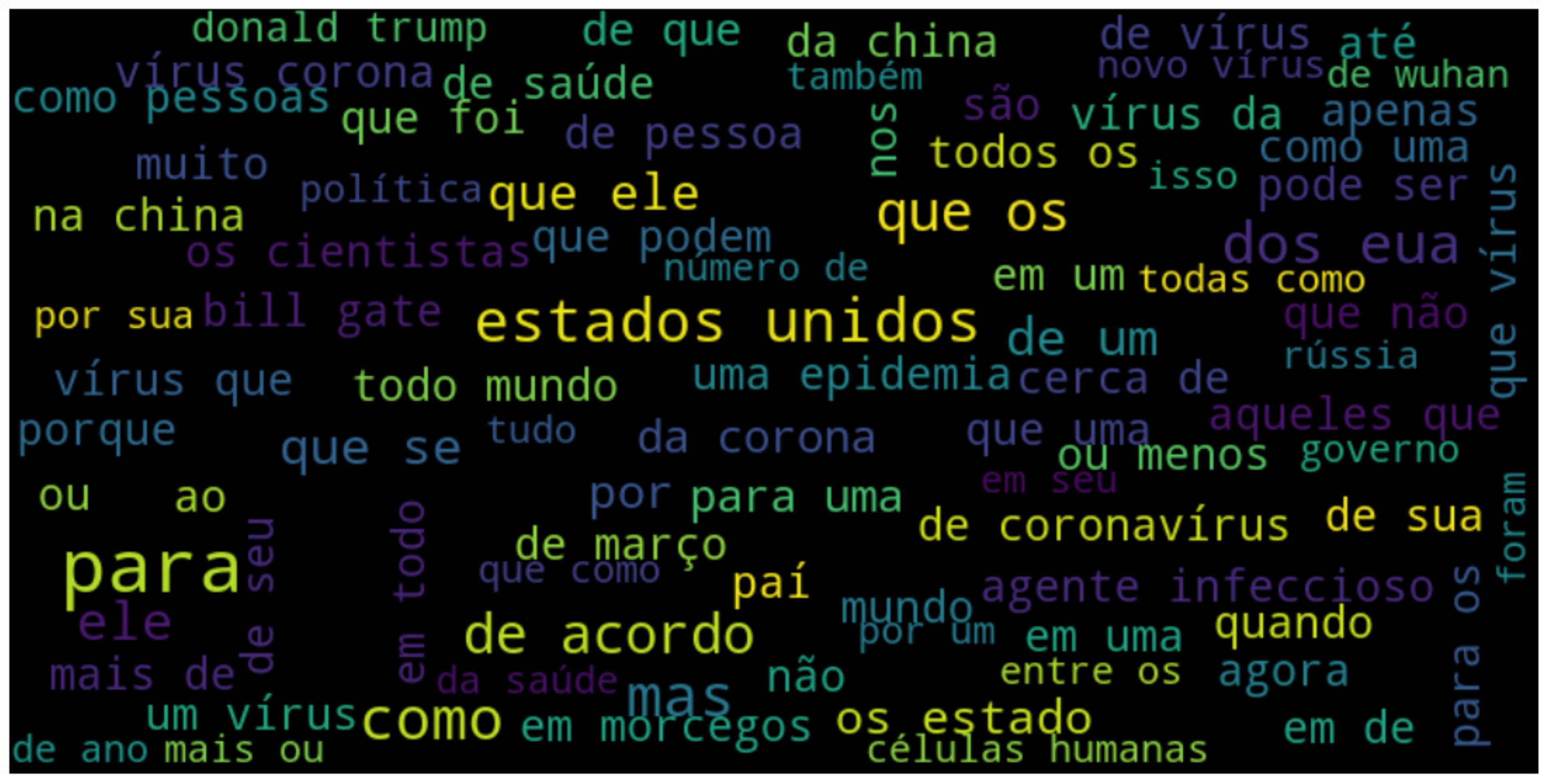}}
      \caption{\textit{Word clouds} after \textit{data augmentation}.} {Source: The Author}
      \label{steady_state}
\end{figure}
In both Figures 5 (a) and (b), it is possible to identify that there was significant growth in the most common terms presented in the \textit{word cloud} of the \textit{datasets}. It is also possible to notice that Figure 5 (b) in particular, \textit{word cloud} from ``\textit{Fake Train}", is a little more crowded with words than Figure 5 (a). So, it can be indicated that the \textit{data augmentation} process impacted significant changes in both datasets.

\subsection{Modeling and Training} 
An architecture based on the BERT classification model was used for the training phase. This model has been widely used for text classification because of its success in several architectural categories. According to BERT's creators \cite{DBLP:journals/corr/abs-1810-04805} this model is pre-trained on a large body of text and then fine-tuned for use in specific tasks.

For this work, in addition to BERT, \textit{Droupout}\footnote{\textit{Dropout} techniques were used: Technique to reduce complex co-adaptations of neurons, forcing a neuron to learn more robust features that are useful in conjunction with many random subsets of other neurons.} and \textit{Dense Layer}\footnote{\textit{Dense Layer}: It is a layer in which each input neuron is connected to the output neuron, and the parameter units only inform the dimension of your output} with five layers. All training takes place in 10 epochs\footnote{Training epoch: refers to a training cycle with the entire \textit{dataset}. Typically, training a neural network takes more than a few epochs.} and three epochs of patience. Patience is an indicator referring to the number of epochs with no improvement in accuracy before training stops. If there is no improvement in one season and the next one has a negative performance, the training stops. This prevents \textit{LOSS} from growing and maintains accuracy at its best.


\begin{figure}[hbt!]
\centering
\includegraphics[width=.7\textwidth]{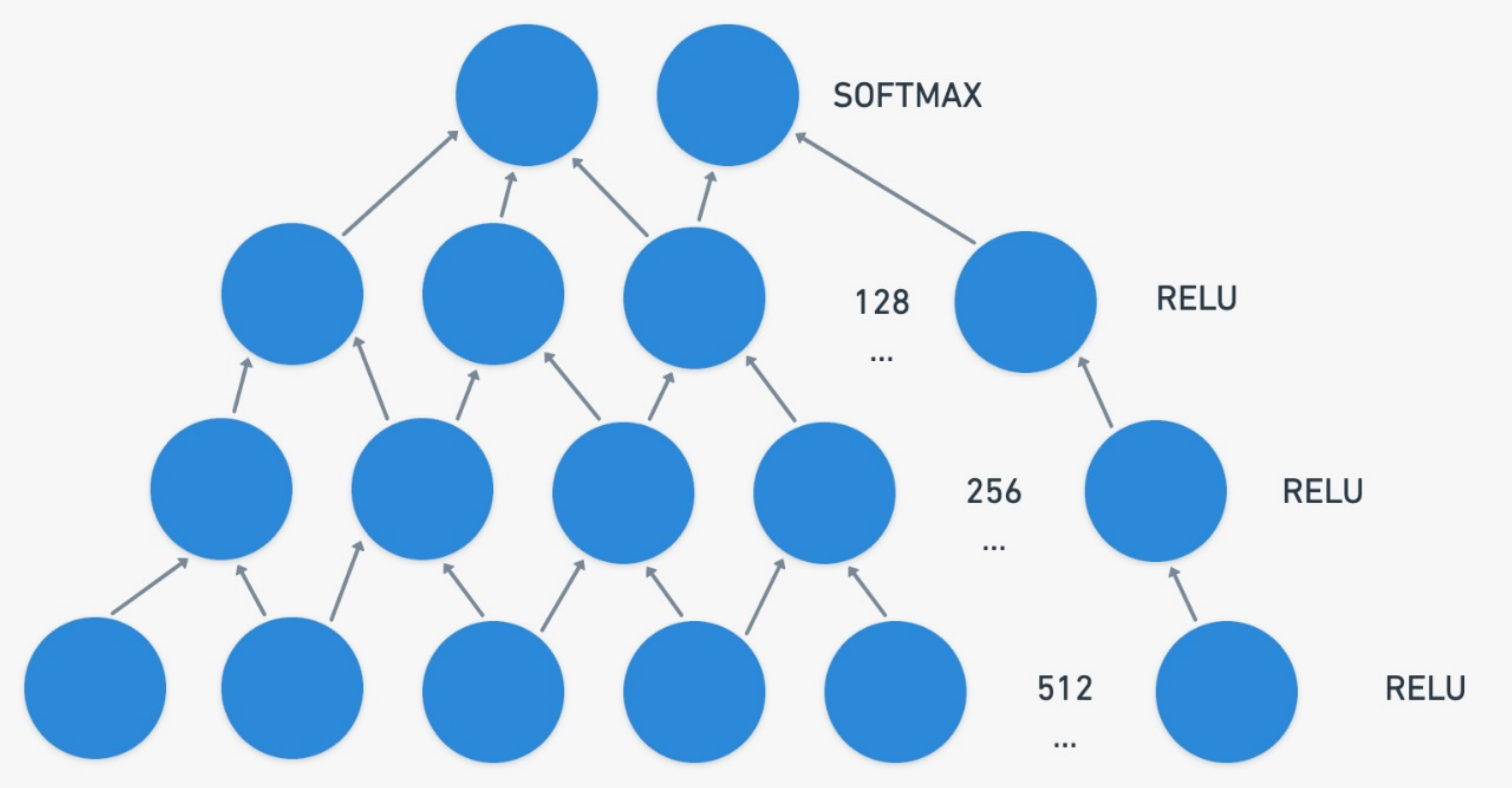}
\caption{Neural network neurons architecture.}{Source: The Author}
\label{}
\end{figure}

Figure 6 represents how the structure of neurons is designed. As the BERT algorithm can only accept sentences of up to 512 words, it is necessary to preprocess the data (long news) to feed the algorithm. The author follows this idea and segments each part of the text into multiple subtexts of a maximum of 150 words. Subtexts will be overlapped where, specifically, the last 30 words of the first subtext will be the first 30 words of the second subtext.

Then the original text is preprocessed into input resources that BERT can read. The process consists of tokenizing and converting the original text into token IDs, so the algorithm can read them. Words are tokenized based on the vocabulary dictionary they were pre-trained on (about 30,000 words), and unfamiliar words are broken down into smaller words contained in the dictionary. The maximum string length is also specified to fill all strings with the same length \footnote{The final string length would be longer than specified, as the BERT tokenizer will split unknown words into several known small words.}.

After all the texts are prepared, we move on to training the neural network using the translated and augmented \textit{datasets}. The training structure made by the author used five epochs. However, as the \textit{datasets} had their size almost doubled, the training was changed to 10 epochs so that there is no loss of pro efficiency in case of a short training time limit.

\begin{table}[hbt!]
\centering
\resizebox{\textwidth}{!}{%
\begin{tabular}{l|c|c|}
\cline{2-3}
                                & \multicolumn{1}{l|}{BEFORE DATA AUGMENTATION} & \multicolumn{1}{l|}{AFTER DATA AUGMENTATION} \\ \hline
\multicolumn{1}{|l|}{VALIDATION LOSS} & 0.3638 & 0.3297 \\ \hline
\multicolumn{1}{|l|}{VALIDATION ACCURACY} & 0.9487 & 0.9172 \\ \hline
\multicolumn{1}{|l|}{GENERAL ACCURACY} & 0.9516 & 0.9913 \\ \hline
\multicolumn{1}{|l|}{LOSS} & 0.3611 & 0.0293 \\ \hline
\end{tabular}%
}
\caption{Comparative table indicating the values before and after applying the technique.}{Source: The Author}
\end{table}

The data plotted in Table 1 indicate that there was a significant increase in training accuracy and a decrease in the \textit{loss} percentage, also showing a significant drop in validation accuracy, one of the indicators that correspond to real data compared to the predictions.

\subsection{Validation and Results}

After training, precision was used as the primary metric for model validation, in addition to comparing values prior to the \textit{data augmentation} process, to analyze whether there was any change in three essential parameters: ``Training Accuracy ", `` Accuracy of Validation, "~and ``Accuracy of Test", where each item represents the precision of each step (``training", ``test"~and ``validation").

\begin{table}[hbt!]
\centering
\resizebox{\textwidth}{!}{%
\begin{tabular}{l|c|c|}
\cline{2-3}
                                      & \multicolumn{1}{l|}{BEFORE DATA AUGMENTATION} & \multicolumn{1}{l|}{AFTER DATA AUGMENTATION} \\ \hline
\multicolumn{1}{|l|}{TRAINING ACCURACY} & 0.9596157073974609 & 0.9944672584533691 \\ \hline
\multicolumn{1}{|l|}{VALIDATION ACCURACY} & 0.954023003578186 & 0.9166180491447449 \\ \hline
\multicolumn{1}{|l|}{TEST ACCURACY} & 0.9438775777816772 & 0.9254571199417114 \\ \hline
\end{tabular}%
}
\caption{ Comparative table indicating the values before and after applying the technique}{Source: The Author}
\end{table}
It can be seen in Table 2 that there was a drop in the values of "Validation Accuracy" and "Test Accuracy" after the application of \textit{data augmentation}.~Despite the considerable increase in the "Training Accuracy" indicator, the drop in other indicators may be associated with the translation of the news into Portuguese. In any case, the post \textit{data augmentation} values remained firmly above 91 percent.

For validation and in order to verify hit rates in the news that the neural network has never seen before, a case study was approached with the following metrics:

\begin{enumerate}
    \item Analyze 20 real news and 20 fake news, both taken from the \textit{datasets} ``\textit{Real Test}"~and ``\textit{Fake Test}".
    \item Analyze 20 real news and 20 fake news, both taken from genuine sites and posts in Brazilian territory.
\end{enumerate}

First validation: in the test with the \textit{dataset} ``\textit{Fake Test}", the neural network got 95 percent correct (19 out of 20 news), in addition to significant improvements in respect of accuracy. In the test with the news real, the neural network obtained 95\% accuracy (19 of the 20 news), showing a significant improvement in accuracy values.

Second validation: in the test with real news, the neural network was 95\% correct (19 out of 20 news), showing good stability in accuracy between the news. In the test with fake news, the neural network was 70 percent correct (14 out of 20 news). Looking at the last test of this validation step, it was decided to generate a \textit{word cloud} with the most common words to understand why the neural network made a mistake in certain news.

\begin{figure}[hbt!]
\centering
\includegraphics[width=.7\textwidth]{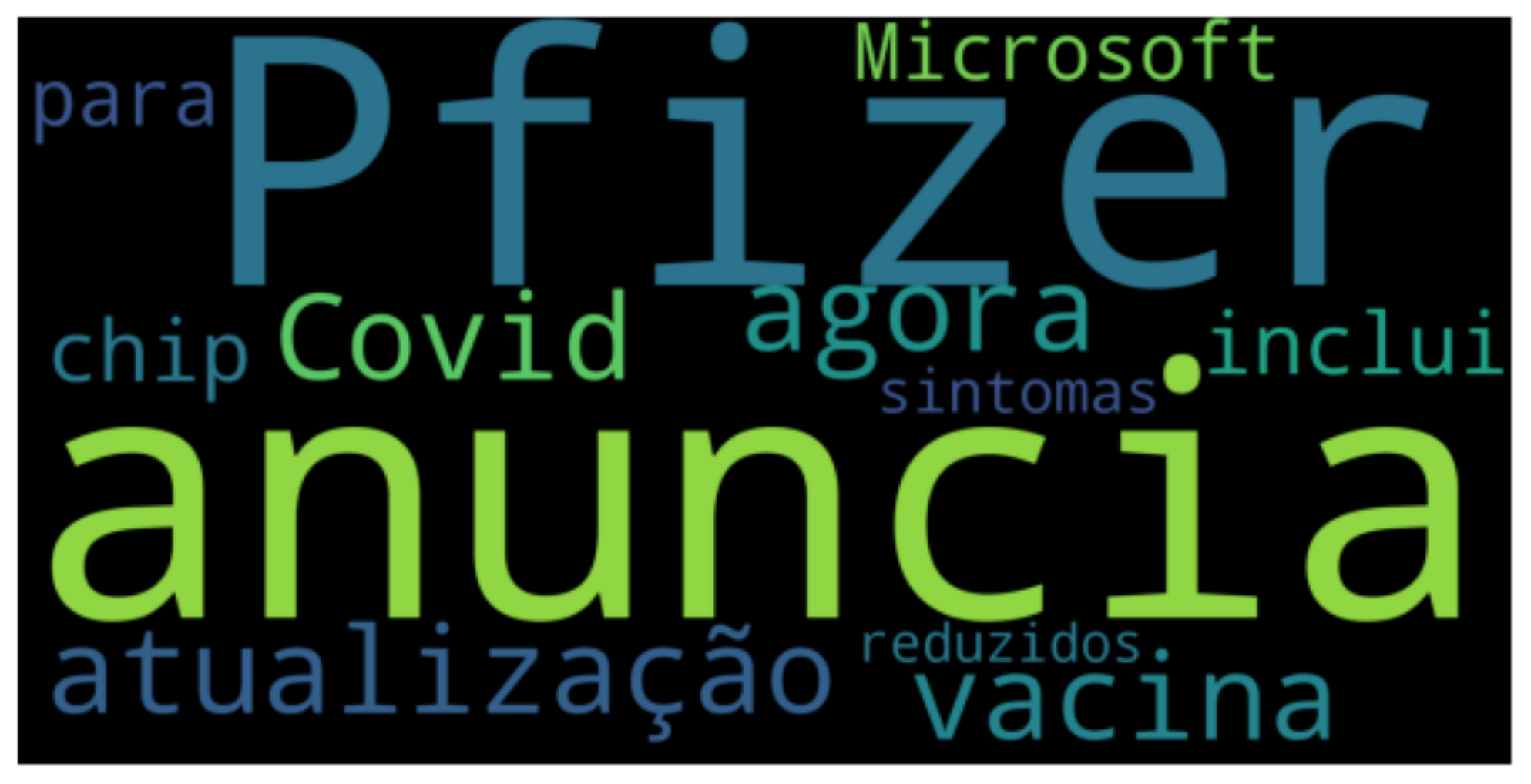}
\caption{Most common words in the news are misclassified.}{Source: The Author}
\label{}
\end{figure}
In Figure 7, it is possible to notice that words like ``Pfizer", ``\textit{chip}"~and ``Microsoft"~appear as the most common keywords among the sentences where the neural network applied a wrong classification. Because, in \textit{the dataset}, there is no consistent information containing these words (the closest to ``\textit{Microsoft}"~ would be ``Bill Gates", but the neural network has not been trained to relate terms). As the \textit{dataset} is of foreign origin, it is possible that, in other countries, fake news is created with some different characteristics from Brazilians ones, and adapting to the Brazilian context is a real challenge.

It is also possible to explain the case of the first validation: both analyzed news came from \textit{datasets} that were also translated. This may explain the percentage difference between the two validation steps. In addition, even before the translation of the \textit{datasets} the AI was already classifying false positives in tests, information that can be considered through the percentages of accuracy shown in Tables 1 and 2

All this data can be consulted on GitHub\footnote{Validation test: https://github.com/mauriciosena/tccdataaugmentation/tree/main/Tests}

\section{Future works}
In the course of this work, some functions emerged that could help in the AI training process, as well as in the final classification accuracy:
\begin{enumerate}
    \item Correlational analysis of words: as it was possible to notice in the second stage of validation of new classification, the neural network could not relate the word ``Microsoft"~ with ``Bill Gates"~ because it was not trained to do so. This practice can be implemented during the \textit{data augmentation} process as a way of looking for relationships between words in the same context (not just by synonyms) and building an alternative \textit{dataset} so that the neural network can classify news with greater accuracy.
    \item Addition of Brazilian news to \textit{datasets}: real or fake, adding more messages in the Brazilian context, not only translating foreign news, can have a significant impact on the neural network during the classification of national news.
    \item Classification of the type of fake news: implement a structure where the neural network can identify beyond just real or fake news, being able to sub classify fake news as misleading, partially false, unsustainable (when there is no statistical evidence to prove it, for example), and other kinds of frauds.
\end{enumerate}.

\section{Conclusions}
As the tests demonstrate, the \textit{data augmentation} was effective in increasing the accuracy of the analyses. It was possible to detect that the complete translation of a \textit{dataset} into another language implies a loss of precision when applied in the context of the country. However, with the translation combined with the \textit{data augmentation} technique, it was possible to use the neural network to analyze true and false news in the Brazilian language, presenting a very significant percentage of accuracy.

On the other hand, it is understandable to note that there was a considerable drop in detecting fake news in the Brazilian language, which the neural network never had contact with concerning the news in the dataset. This is because all the training was based on a translated \textit{dataset}, where the sentences and contexts, even efficiently translated, generally did not correspond to real situations in Brazil.

However, this research project proved to be possible, improving existing functionalities, and implementing new functions in an already developed project, presenting a significant increase in the margin of precision. The entire project is available on GitHub\footnote{Project GitHub: https://github.com/mauriciosena/}.


\bibliographystyle{unsrt}  
\bibliography{references}  

\end{document}